\documentclass[sigconf]{acmart}

\usepackage{booktabs} 

\setcopyright{rightsretained}

\acmYear{2018} 
\setcopyright{acmcopyright}
\acmConference[GECCO '18 Companion]{Genetic and Evolutionary Computation Conference Companion}{July 15--19, 2018}{Kyoto, Japan}
\acmBooktitle{GECCO '18 Companion: Genetic and Evolutionary Computation Conference Companion, July 15--19, 2018, Kyoto, Japan}
\acmPrice{15.00}
\acmDOI{10.1145/3205651.3208300}
\acmISBN{978-1-4503-5764-7/18/07}

\editor{Jennifer B. Sartor}
\editor{Theo D'Hondt}
\editor{Wolfgang De Meuter}

\author{Rahul Shivnarayan Mishra}
\affiliation{%
  \institution{Indian Institute of Technology Guwahati}
}
\email{rahulm@nvidia.com}

\author{Tushar Semwal}
\affiliation{%
  \institution{Indian Institute of Technology Guwahati}
}
\email{t.semwal@iitg.ac.in}

\author{Shivashankar B. Nair}
\affiliation{%
  \institution{Indian Institute of Technology Guwahati}
}
\email{sbnair@iitg.ac.in}

\usepackage{footmisc,hyperref}
\usepackage{verbatim}

\renewcommand{\thefootnote}{\fnsymbol{footnote}}

\usepackage{enumerate} 

\usepackage[caption=false,font=footnotesize]{subfig}

\usepackage{stfloats}

\usepackage{algorithm,algorithmic}
\usepackage[normalem]{ulem}
\usepackage{fontawesome}


\begin{document}
\title{A Distributed Epigenetic Shape Formation and Regeneration Algorithm  for a Swarm of Robots}

\begin{abstract}
Living cells exhibit both growth and regeneration of body tissues. Epigenetic Tracking  (ET), models this growth and regenerative qualities of living cells and has been used to generate complex 2D and 3D shapes. In this paper, we present an ET based algorithm that aids a swarm of identically-programmed robots to form arbitrary shapes and regenerate them when cut. The algorithm works in a distributed manner using only local interactions and computations without any central control and aids the robots to form the shape in a triangular lattice structure. In case of damage or splitting of the shape, it helps each set of the remaining robots to regenerate and position themselves to build scaled down versions of the original shape. The paper presents the shapes formed and regenerated by the algorithm using the Kilombo simulator.\end{abstract}


\maketitle
\section{Introduction}
Almost all biological beings are aware of their shape and size. Many researchers have strived to mimic these complex physical and cognitive living entities in the form of a Single Robot Systems (SRS). An SRS is aesthetically and physically closer to these beings but is indeed a highly sophisticated system. Simple small robots with insufficient capabilities may not be able to perform complex tasks that an SRS could perform, but a collection of a large number of simpler robots having limited capabilities which cooperate to fulfil a goal might be able to perform as effective and efficient as an SRS. The goal could be some form of self-organization such as to maintain a particular shape formation . If we consider each robot in such a swarm to be the basic unit of a body, then the swarm metaphorize the biological being. A swarm of robots collectively executing a task can be more robust, adaptive and efficient as compared to the SRS. Swarms play a significant role in applications such as exploration of inaccessible areas  and mapping \cite{schmickl2006collective,correll2006collective}, hazardous tasks executions \cite{kumar2003cognitive}, pattern formation \cite{rubenstein2014programmable}, collective transportation \cite{gross2006transport}, etc.

Algorithms for shape formation could be classified into those that use leader/neighbour-following methods, potential field based ones and those that are nature-inspired.  In the first type \cite{chen2009backstepping,choi2007fuel,kanjanawanishkul2008model}, the primary goal is to follow the leader which in turn has all the information required to make the moves. The robots need to arrive at a consensus and the notion of a leader gives the overall system a centralized flavour. However, these works constitute rigorous theoretical proofs which makes them relatively complex \cite{meng2013morphogenetic}. In the potential field based methods \cite{krishnanand2005formations,masoud2007decentralized,hsieh2008decentralized}, the robots follow potential field gradients which constitute the resultant of virtual attractive and repulsive forces. Though successfully implemented in several scenarios, the robots may get stuck at local optimal points \cite{meng2013morphogenetic}.  Nature-inspired shape formation algorithms are based on the manner in which swarms function in an inherently distributed and decentralized manner. A swarm can be made to achieve a goal using either a centralized or a decentralized approach. The former may seem simple and faster but is not always an appropriate solution since the swarm is dependent on a single controlling entity. If this entity fails, the entire system can collapse. Further, centralized control is difficult and expensive to scale. A decentralized approach seems much more efficient and robust. Nature solves a gamut of problems using such decentralized and distributed approaches. Researchers have thus proposed several bio-inspired algorithms.

Cheng et al. \cite{cheng2005robust} describe a pheromone and flocking inspired gas expansion model for coordinating a swarm of identically programmed robots to spatially self-aggregate into arbitrary shapes using only local interactions. Rubinstein et al. \cite{rubenstein2014programmable} have proposed a self-assembly algorithm to build shapes using a swarm of Kilobots robots that have limited capabilities. Their algorithm does not support regeneration and thus cannot evolve to form the same shape when some portion of it is cut off. The improvised version of this work is described in \cite{rubenstein2009scalable} wherein the authors augment a self-repairing feature. The repair of the mutilated shape, however, requires the human being to intervene and provide the number of remaining robots in the swarm, thus making the system semi-autonomous. George et al. \cite{george2003biological} describe a cell-based programming model capable of creating symmetrical shapes such as cubes, spheres, etc. using numerous cells. If the shape were to be cut or damaged, the cells within detect the same and regenerate or self-heal the destroyed portion. Apart from these, a few bio-inspired algorithms based on hormones \cite{shen2004hormone,rubenstein2009regenerative,rubenstein2010automatic} and cellular mechanisms \cite{mamei2004experiments,taylor2007pattern} have also been reported. Gene regulatory networks have also been employed for shape formation in changing environments \cite{meng2013morphogenetic,jin2012hierarchical,taylor2007pattern}. Though they augment their work with real-robot experiments, the differential equations used to implement the kinematics involved can make the system complex.

An algorithm for a set of robots to generate and regenerate shapes needs to be simple, distributed and decentralized, autonomous and be capable of handling variable-sized complex shapes. Nature-inspired algorithms provide for comparatively simple mechanisms with lower computational overheads \cite{yang2016nature}. They are inherently distributed and decentralized and hence have an edge over other conventional methods for shape formation and regeneration.

Epigenetic Tracking (ET), first introduced by Fontana \cite{fontana2010epigenetic}, is one such algorithm which models the growth and regenerative qualities exhibited by a colony of cells found in a living body. ET is defined as the method to generate arbitrary 2D and 3D shapes by using evolutionary techniques. It explains the process by which a seed grows into a whole living being. In nature, we can find certain peculiar species which have a peculiar tendency to grow their body parts. For example, a lizard regrows its tail after it sheds it off, while a starfish regenerates its leg if it is cut off. Human babies when inside the mother's womb, are capable of regenerating their limbs. All this regeneration is done by the body cells themselves without the requirement of any central system. Each cell contains a gene which decides the shape of a being. A cell divides and replicates itself as per the gene within the cell. A gene is similar to an instruction manual which aids cells to reproduce and form arbitrary shapes.

In this paper, a distributed novel Epigenetic Tracking based shape formation and regeneration algorithm has been proposed for a swarm of robots. Each robot is modelled in the form of a cell containing a gene. The algorithm is tested on Kilobots \cite{rubenstein2014programmable} which are small robots with limited computation and wireless communication capabilities.  Experimental results performed using a simulator have been presented. To the best of our knowledge, this is the first occurrence of a work where ET has been used for shape formation and regeneration in a swarm of robots. 

Subsequent sections of this paper describe the shape formation and regeneration problem, the algorithm and associated terminologies and the results and conclusions.

\section{Shape Formation and Regeneration}
In this section, we have discussed the problem of shape formation and regeneration by a swarm of robots using a decentralized algorithm. The same is followed by a detailed description of the various terms and methods used in the algorithm.

\subsection{Problem Definition}

This paper focuses on the problem of shape formation and shape regeneration exhibited by a swarm of robots without a central controller. All the robots are aware of the target shape, and they work collectively to form a given shape and is capable of detecting any damage done to the shape. The damage could be in the form of removal or failure of robots. After detecting the damage, the remaining robots work collectively to form a smaller scaled version of the specified shape. The primary objective is thus to use only local interactions and computations onboard the robots to once again reorganize, assemble and regenerate the original shape with the remaining robots in the swarm.

We propose the use of an Epigenetic Tracking based algorithm capable of running on small, low-cost robotic platforms such as Kilobots \cite{rubenstein2014programmable}. Kilobot is one such robotic platform which has limited computational capabilities and is capable of achieving short-range omnidirectional communication with their neighbours via infrared rays. They are devoid of any other sensor. The algorithm is state-based, meaning that a robot is always in a unique state. A robot thus performs an action(s) based on its current state and the information obtained from the neighbouring robots.

\begin{figure}[t]
    \centering
    \includegraphics[scale=.6]{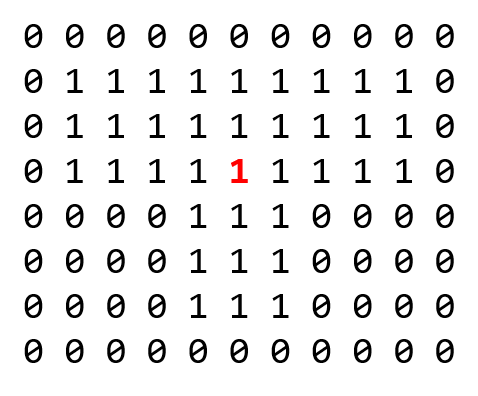}
    \caption{A sample input bitmap image of shape `T'} 
    \label{input}
\end{figure}

\subsection{Preliminaries}
Epigenetic Tracking (ET) models the biological development process found in living cells. Coupled with evolutionary techniques, the model is capable of generating complex 2D or 3D shapes from artificial cells. In this model, an entire shape emerges from a single cell through a sequence of two actions,  \textit{mitosis} and \textit{apoptosis}. Mitosis is the process of cell division where a cell produces a clone of itself. Apoptosis is programmed cell death, which keeps control of cell population. The two actions are driven and directed by a gene. A gene is a long strand of encoded information used by a cell to decide whether it has to perform mitosis or apoptosis. After a series of cell actions, the desired target shape emerges from a single cell. A gene is similar to an instruction manual which guides the process for the shape formation.

Inspired by this simple distributed and robust shape formation techniques modelled by ET, we propose a biologically inspired algorithm for shape formation and regeneration exhibited by a swarm of robots. Each robot has a unique ID and is modelled as a cell containing a gene which has encoded information about the target shape. The algorithm takes the input in the form of a binary image with given rows and columns while the output is a swarm of robots positioned in the manner similar to the image. A sample input in the form of a binary image to form a `T' shape is shown in Fig. \ref{input}. Since the pixel values are either \textit{0} or \textit{1}, the image will be referred to as a bitmap array where \textit{1} denotes the presence of a robot which is part of the shape while a \textit{0} represents a void space. The number of \textit{1}s indicate the consensus of robots required to form the shape, and thus each robot initially knows about the current swarm population.

Given an input binary image, the gene required for the shape formation is computed and transferred to all the robots constituting the swarm. The robots (or cells) then use this gene to form a shape in a distributed and decentralized manner. In this work, we have taken the coordinates of the central pixel of the input bitmap (red coloured \textit{1} in Fig. \ref{input}) as the reference or origin. The position of the remaining pixels is described and used with reference to this origin. It may be noted that coordinates of any pixel could be chosen as the origin. 

\renewcommand{\thefootnote}{\faFileText}

\subsubsection{Skew Coordinate System}

Though the use of the Cartesian arrangement shown in Fig. \ref{input} is straightforward, it suffers from the fact that all the neighbours of a \textit{1} are not equidistant. The \textit{1}s at the diagonal points are at a farther distance than the \textit{1}s at the edges. This can lead to instability especially while real robots are used to form shape. Before computing the gene, the input image is thus, skewed in a manner that every robot is equidistant from the other.  We followed a simple process to generate the skewed image. Consider the input shape in Fig. \ref{scs}. The top
row of the image is first shifted left by half the distance between the two pixels (or two adjacent \textit{1}s or \textit{0}s in the row).  We then shift this row leftwards along with the one just below it together by another half distance. 
In the next step, the top row and the two rows immediately below it are shifted leftwards together, by the same distance. This shifting process is repeated till eventually, the whole
image gets skewed to form a triangular lattice structure. 

The computation steps involved in the making of the gene are described in the following subsections.

\subsubsection{Gene Computation}
In the current context, the metaphor gene is a collection of 1-Dimensional ($D$) vectors each with three fields of information -- a \textit{Tag} (Tg) which denotes a pair of $X$ and $Y$ location coordinates, a \textit{Flag} (F) and the number of \textit{Nearest Neighbors} (NN) of a robot at the corresponding (X,Y) location. 
To compute a gene from an input binary image, the $Tg$, the $F$ and the \textit{NN} fields of information for each pixel value in the input image is calculated and concatenated to create a table of 1-$D$ vectors. This table forms the gene for a given shape. The significance of each of the constituents of these vectors is presented below:

\begin{figure}[t]
    \centering
    \includegraphics[scale=.3]{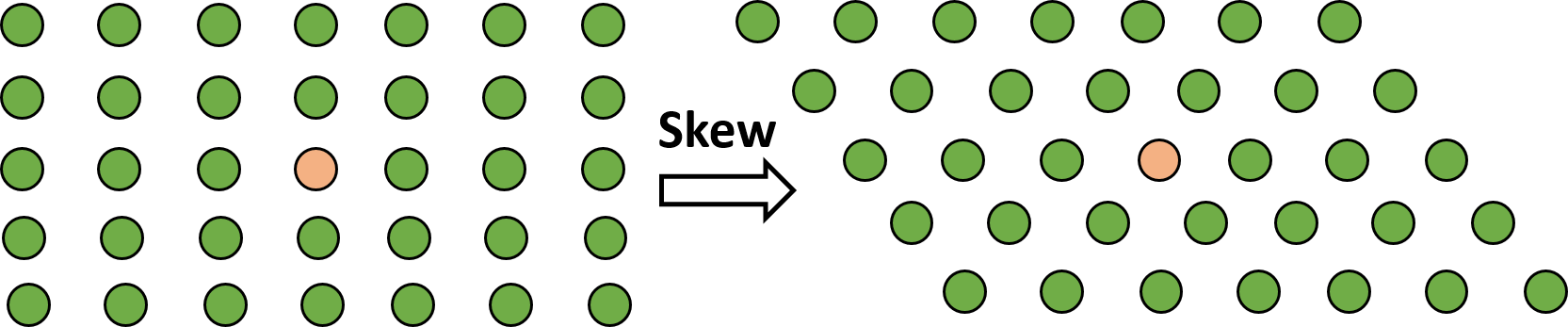}
    \caption{Transformation of arrangement used in the proposed algorithm
        } 
    \label{scs}
\end{figure}

\begin{enumerate}
\item {Tag ($Tg$)}: Tag is the (X,Y) pixel coordinate  of the input bitmap. For e.g., $Tg$ for the origin is (0,0). It aids the robot in determining its position within the swarm. 

\item Flag ($F$): If a bit at a coordinate (X,Y) is equal to 1, the respective flag is set to 1 else it is set to 0. The flag represents whether a robot at a position (X,Y) is inside the shape (F = 1) or outside the shape (F = 0). 


\item Number of Nearest neighbors (\textit{NN}): This is a value ranging from 0 to 6 which indicates the number of nearest neighbors a robot at a position (X,Y) is allowed to have. The NN value for a bit at location (i,j) in the input bitmap image is equal to the number of \textit{1}s at the location (i-1,j), (i-1,j+1), (i,j+1), (i+1,j), (i+1,j-1) and (i,j-1). 

\end{enumerate}

\begin{figure}[t]
    \centering
    \subfloat[]{\includegraphics[scale=0.7]{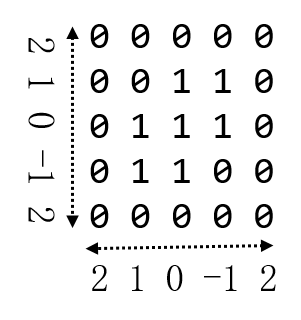} \label{bitmap}}
    \hfill
    \subfloat[]{\includegraphics[scale=.6]{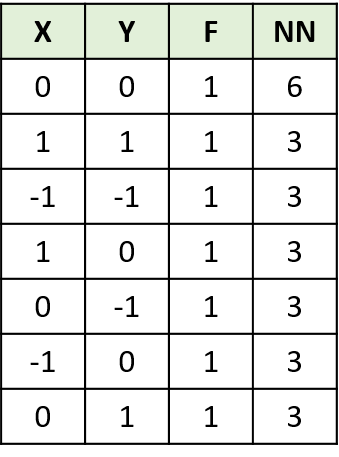} \label{gene}} 
    \caption{(a) A skewed input bitmap image of an arbitrary shape (b) Computed gene}
    \label{genetable}
    
\end{figure}

Fig. \ref{bitmap} and \ref{gene} shows an input bitmap image and the computed gene, respectively. As an example, the (X,Y), $F$ and the \textit{NN} values for the bit at the centre of the bitmap image shown in Fig. \ref{bitmap} are (0,0), 1 and 6, respectively.

A robot at a location (X,Y) is aware of the count of closest neighbours it is supposed to have by looking into the gene with the tag (X,Y). These neighbours should be at an equal distance say a constant $d$ so that the robot can differentiate them from the rest of the swarm. Since all the robots maintain a distance of $d$ from each other, they form a triangular lattice structure as shown in Fig. \ref{scs}. After forming the shape, the robot keeps monitoring this count. If it decreases, the robot discerns that formed shape is harmed. Thus, damage detection forms the primary aim of triangular lattice arrangement. The sole purpose of the SCS was to have a coordinate system which allows unique referencing of robots in the triangular lattice arrangement. 

\section{Self-organization and Self-regeneration}
In this section, we propose an ET based distributed and decentralized algorithm for shape formation and regeneration in a swarm of robots.

\subsection{Epigenetic Tracking based Algorithm}
The overall process of shape formation and regeneration exhibited by a swarm of robots using the proposed algorithm, follows a Discrete Finite Automaton (DFA) with eight states namely, $Queued$, $Search$, $Inactive$, $Active$, $Quasi$, $Stable$, $Danger$ and $Leader$. Table \ref{tab1} provides an abstract description of each of these states. Since the process of regeneration is after shape formation and for the sake of clarity, the algorithms for both the processes, i.e., formation and regeneration has been explained separately using the states described in Table \ref{tab1}.

We also introduce a gradient value termed the Timestep (TS) in the proposed algorithm. Every robot which has been placed in the shape has a gradient value TS to assist robots in reaching \textit{Stable} state according to their order of arrival. Robots in the \textit{Search} states who are trying to determine their new locations during the shape formation process always try to locate themselves near a robot in the \textit{Active} state with minimum TS value compared to their other neighbours in \textit{Active} state. Consider two robots $R_A$ and $R_B$ in the \textit{Active} state with TS values 4 and 5 respectively. This indicates that the \textit{NN} criteria for $R_A$ needs to be fulfilled first as $R_A$ was placed into the shape/cluster before $R_B$.  The TS thus prevents the formation of holes or gaps in the shape which would then cause the shape to become irregular at the end.

\begin{table}[t]
    
    \caption{States and their description}
    \begin{tabular}[width=\linewidth]{|p{0.85cm}|p{7cm}|}
        \hline
        \textbf{State} & \textbf{Description} \\
        \hline
        \textit{Queued} & All robots are initially in a swarm repository in a connected form. Using broadcasts, these robots find the one having the lowest ID. This robot changes its state to \textit{Search} and moves out of the repository towards the seed robots to localize itself. The robot with the next lowermost ID moves out after a predetermined amount of time. \\
        \hline
        \textit{Search} & Robots in this state search for a location within the shape by moving along the periphery of the currently formed shape. After finding an appropriate location, the robot places itself and transits to the \textit{Inactive} state. \\
        \hline
        \textit{Inactive} & 
        In this state, the robot continuously monitors all of its nearest neighbours which are in the \textit{Active} state. As soon as a neighbour in an \textit{Active} state transits to the \textit{Stable} state, the robot in the \textit{Inactive} state transits to the \textit{Active} state. This transition avoids the chances of gaps/holes appearing within the shape. \\
        \hline
        \textit{Active} & 
        A robot remains in this state till the NN value stored within it matches the actual number of neighbours. When these two values are equal, the robot transits to the \textit{Stable} state. If a robot in the \textit{Active} state notices one of its nearest neighbours to be in the same state but having a TS value less than its own, it transits to the \textit{Quasi} state. \\
        \hline
        \textit{Quasi} & 
        A robot in the \textit{Quasi} state transits to the \textit{Active} state after a predetermined time. \\
        \hline
        \textit{Stable} & 
        Robots in this state form part of the final shape. These robots continuously count their neighbours to ensure that the shape is maintained. If they detect the absence of any, they transit to the \textit{Danger} state. \\
        \hline
        \textit{Danger} & 
        Robots in this state participate in the process of leader election. The robot with the minimum ID amongst these robots is elected the leader which in turn transits to the  \textit{Leader} state. \\
        \hline
        \textit{Leader} & The robot in this state estimates the number of remaining connected robots in the fragment of the shape, generates a scaled-down version of the given shape (gene) and communicates the same with the remaining connected robots. Finally, it selects two of its neighbours to form the initial three seed robots to start the shape formation once again using the remaining robots. The Leader then transits to the \textit{Active} state. Those from the set of remaining robots whose coordinates match that within the new version of the shape transit to the \textit{Inactive} state while the rest move to the \textit{Queued} state. \\
\hline
     
 \end{tabular}
 \label{tab1}
 \end{table}

 \subsubsection{Shape Formation}
The proposed algorithm for shape formation needs a minimum of three seed robots which need to be placed at the vertex of an equilateral triangle. These seed robots act as an initial frame of reference for the trilateration process executed by other robots to know their location information. The seed robots are respectively initialized with an (X,Y) coordinate, a state, an \textit{NN} value and a TS. The (X,Y) coordinate for one of the seed robots is chosen randomly from the gene. The coordinates of the other two are taken from the appropriate neighbours from the gene. The location of the seed robots decides the position and orientation from where the shape formation will commence. The \textit{NN} values for the corresponding coordinates are also taken from the gene and stored within, by each robot. One of these robots is set to the \textit{Active} state with TS equal to 1, while the other two are initialized to the \textit{Inactive} state with a TS of 2. After the seed robots are initialized, the rest of the robots in the \textit{Queued} state concurrently start to execute the Algorithm \ref{a1} to eventually form the shape represented by the gene. 

Robots in the \textit{Queued} state, one after another, transits to the \textit{Search} state and start searching for a location to occupy around the already placed robots (initially the seed robots). A robot decides on which location to place itself by using the process explained in Sec. \ref{S4.1}. After the robot places itself, it moves to the \textit{Inactive} state if its neighbour count is less the \textit{NN} value of the corresponding tag (X,Y) acquired by the same robot and one of its neighbours is in the \textit{Active} state. The same robot transits to the \textit{Stable} state if its neighbour count becomes equal to the \textit{NN} value. This series of transitions from \textit{Search} state to the \textit{Stable} state is followed by each of the robots in \textit{Queued} state until they all are in the \textit{Stable} state and the shape formation is accomplished.

    \begin{algorithm}[t]
     
        \caption{Algorithm for the shape formation by the swarm of robots}
        \label{a1}
        \begin{algorithmic}[1]
            \WHILE {TRUE}
            \IF {State == \textit{Queued}} 
            \IF {$Id==Minimum$} 
            \STATE State $\leftarrow \textit{Search}$; \COMMENT{the robot has the lowest Id among all the \textit{Queued} robots}
            \ENDIF
            \ENDIF
            
            \IF {State == \textit{Search}} 
            \STATE $MoveTillActiveRobot()$; \COMMENT{Move along the periphery of already formed shape until an \textit{Active} state robot is encountered}
            \STATE Flag $\leftarrow Localization()$; \COMMENT{as per the Sec. \ref{S4.1}}
            \IF {Flag == 1}
            \STATE $Initialize(TS,NN)$;
            \STATE State $\leftarrow \textit{Inactive}$;
            \ENDIF
            \ENDIF
            
            \IF {State == \textit{Inactive}} 
            \IF {Any neighbor move from \textit{Active} to \textit{Stable} state} 
            \STATE State $\leftarrow \textit{Active}$; 
            \ENDIF
            \ENDIF
            
            \IF {State == \textit{Active}}
            \IF {TS of neighbor $<$ TS} 
            \STATE State $\leftarrow \textit{Quasi}$; 
            \ENDIF
            \IF {number of neighbors == NN} 
            \STATE State $\leftarrow \textit{Stable}$; 
            
            \ENDIF
            \ENDIF

            \IF {State == \textit{Quasi}} 
            \STATE $WaitForTime(T)$; \COMMENT{$T$ is a predetermined constant}
            \STATE State $\leftarrow \textit{Active}$
            \ENDIF
            
            \IF {State == \textit{Stable}} 
            \IF {number of neighbors $<$ NN} 
            \STATE State $\leftarrow \textit{Danger}$; 
            \STATE $Algorithm$-$2$();
            \ENDIF
            \IF {New bitmap image received}
            \STATE State $\leftarrow \textit{Queued}$;
            \ENDIF
            \ENDIF
            
            \ENDWHILE
            
        \end{algorithmic}
    \end{algorithm}

\begin{figure}[h]
    \centering
    \includegraphics[scale=.3]{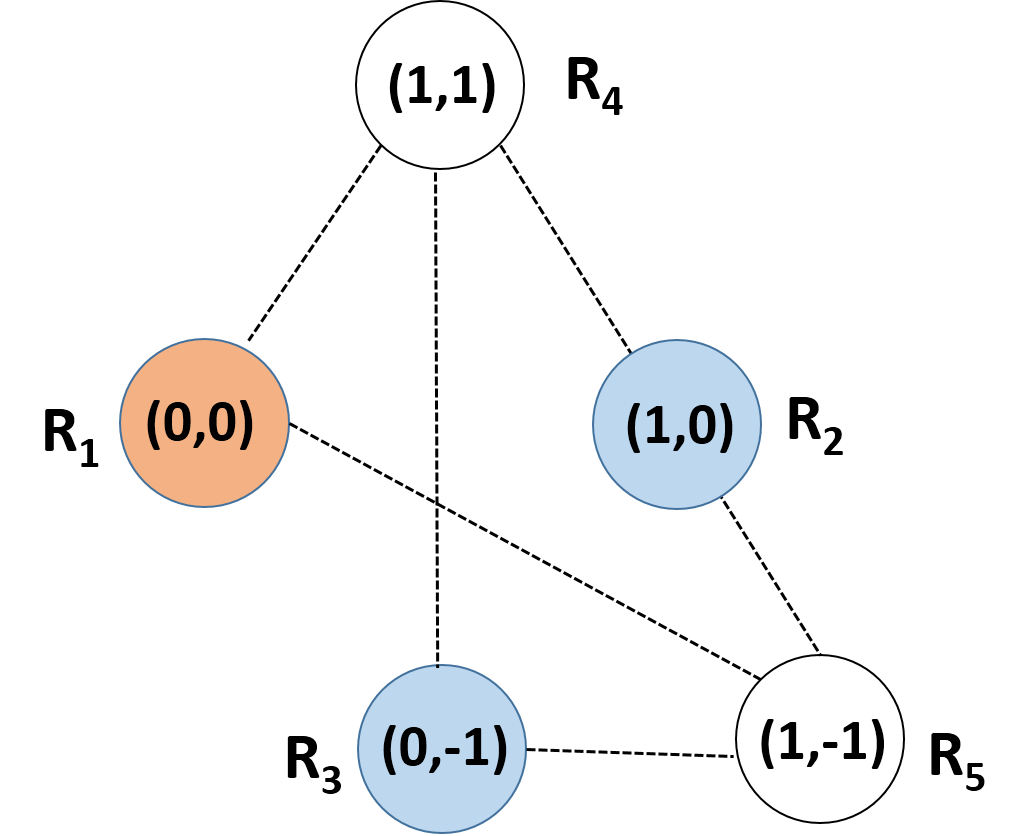}    
    \caption{Position determination by robots} 
    \label{trilat}
\end{figure}

\begin{figure*}[t]
\minipage{0.44\textwidth}
  \includegraphics[width=\linewidth]{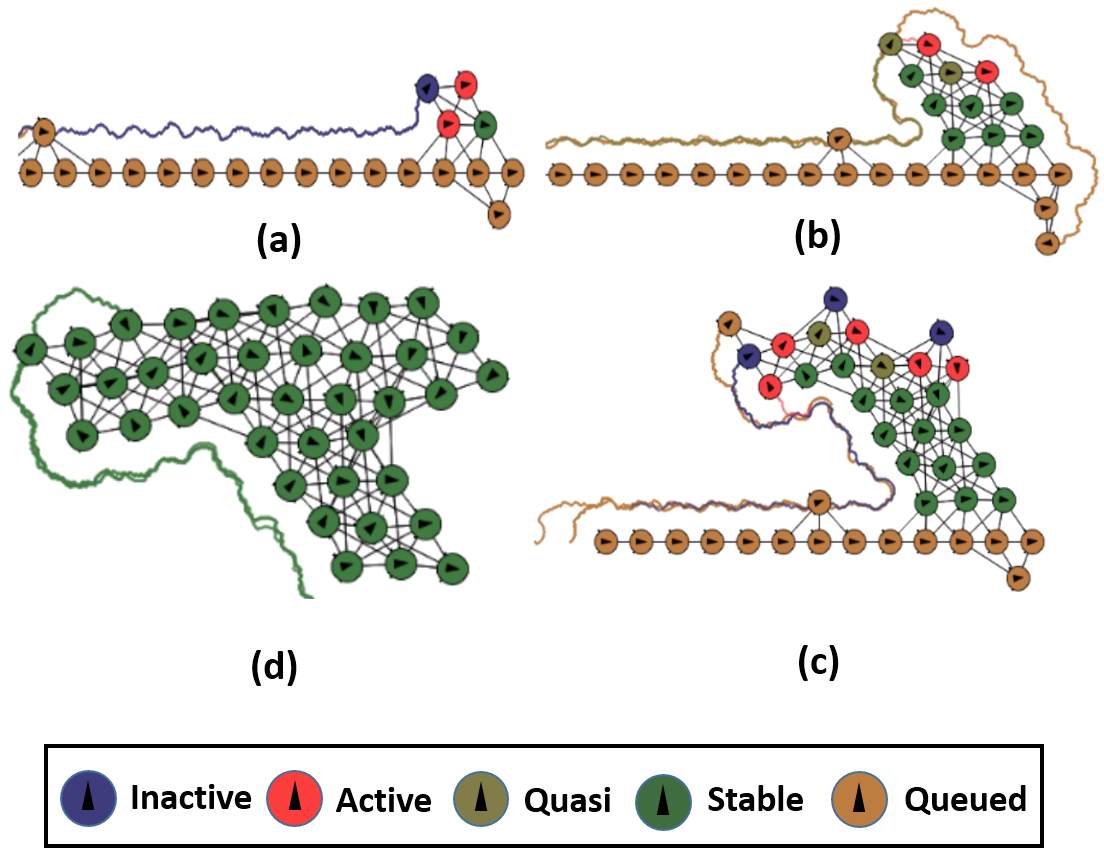}
  \caption{Generation of alphabet `T' in 2 hours 50 minutes and 8 seconds}\label{result_1}
\endminipage\hfill
\minipage{0.35\textwidth}
  \includegraphics[width=\linewidth]{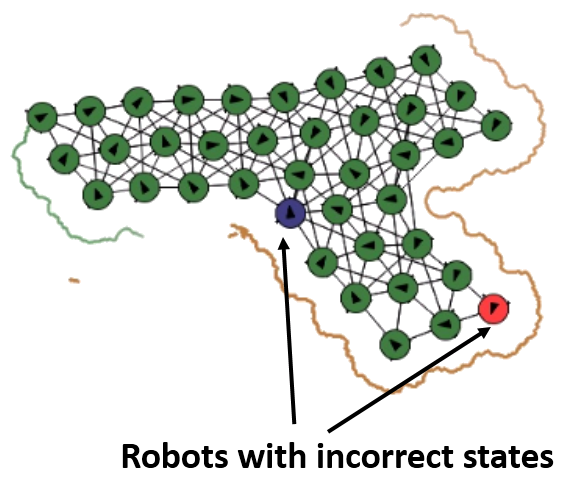}
  \caption{Errors during generation of shape `T'}\label{result_2}
\endminipage
\end{figure*}

\subsubsection{Shape Regeneration}
In practical scenarios, a shape formed by a swarm of robots could get damaged due to various reasons such as malfunctioning of a robot(s), removal of a robot(s), discharge of the batteries, etc. Hence, it is necessary that the same algorithm should facilitate the robots first to detect the damage to the shape and secondly allow them to reorganize to form a scaled-down version of the same shape with the remaining number of robots. If a robot(s) in the \textit{Stable} state finds that its neighbour count has become less than the value NN stored within, it transits to the \textit{Danger} state, constituting the first step in the shape regeneration process. The second step comprises execution of the following sequential phases: 
\begin{enumerate}
\item \textbf{Leader Election}: As soon as the damage is detected, the robots in  \textit{Danger} state compete with those in the same state and elect their leader. We have used the classical method described in \cite{lynch1996distributed} for this leader election, wherein each robot in the \textit{Danger} state broadcasts its ID. Robots in the \textit{Danger} state, which receives an ID lower than their own, back off from the election process and retransmits the lower ID to the others. Eventually, the one which does not receive any ID lower than its own for a certain defined time, transits to the \textit{Leader} state.

\item \textbf{Census}: The robot in the \textit{Leader} state now performs a count of the functional robots in the remnant of the shape it belongs to, by navigating around it and acquiring the $Tg$ of all the robots stationed on the periphery of this shape.  When the \textit{Leader} reaches its original position again, it starts to map these coordinates onto the input bitmap image. By doing so, the Leader obtains the closed area bounded by the remnant shape. By counting the number of \textit{1}s on the boundary of this closed (remnant) shape, the Leader ascertains the number of remaining robots and thereby concludes the census.

\item \textbf{Generation of Scaled Image}: Using the count (say $n$) of the remaining robots in the remnant and the number of robots (say $N$) in the original shape, the Leader estimates the number of robots, say $K$, that have been segregated ($K$=$N$-$n$) and recreate a scaled-down version of the original shape. The leader recreates the scaled image by deleting the \textit{1}s present on the boundary of the current bitmap. The deletion is done sequentially and those \textit{1}s which makes the triangular lattice arrangement unstable are skipped~\footnote{Please refer to the supplementary file for a detailed graphical visualization.}.

\item \textbf{Sharing Scaled Shape and Seeding}: The Leader shares the new scaled down version of the image with all the $n$ robots within the remnant, transforms itself to a seed robot and chooses two of its neighbours as the other two seeds. The remaining $n$ robots become aware of the change in shape and hence transit to the \textit{Queued} state, and the process of shape generation commences again as in Algorithm \ref{a1}.

\end{enumerate}

    \begin{algorithm}[t]

        \caption{Algorithm for the shape regeneration by the swarm of robots}
        \label{a2}
        \begin{algorithmic}[1]
            \WHILE {TRUE}
            \IF {State == \textit{Danger}} 
            \STATE $LId == LeaderElection()$; \COMMENT{the robot has the lowest Id among all the \textit{Queued} robots}
            \IF {$Id == LId$}
            \STATE State $\leftarrow \textit{Leader}$;
            \ELSE
            \STATE State $\leftarrow \textit{Queued}$;\\
            \STATE $Algorithm$-$1$();
            \ENDIF
            
            \ENDIF

            \IF {State == \textit{Leader}} 
            \STATE $CountPopulation()$; \COMMENT{Count the remaining population}
            \STATE $GenerateScaledShape()$; \COMMENT{Scale down the input target shape}
            \STATE Share the new shape with the remaining robot;
            \STATE State $\leftarrow \textit{Active}$;
            \STATE Form new seed robots;\\
            \STATE $Algorithm$-$1$();
            \ENDIF

            \ENDWHILE
            
        \end{algorithmic}
    \end{algorithm}

\subsubsection{Localization within the Shape} 
 \label{S4.1}
During the process of shape formation, the moving robots determine their appropriate location within the shape being formed using the trilateration to determine their position globally. The trilateration method used in this algorithm is different from the conventional one as since we use our own SCS.

The robot herein does not use the formulae in trilateration to determine its (X,Y) coordinates. Instead, it localizes itself by using the coordinates of its neighbours. For instance in Fig. \ref{trilat}, $R_1$, $R_2$ and $R_3$ are the three robots already placed in the shape, each at a distance $d$ from the other. Using SCS, the incoming robot $R_4$, detects $R_1$, $R_2$ and $R_3$ and discovers that it can possibly occupy positions (1,1) or (0,-1). However, since $R_3$ occupying (0,-1) is also in the communication range of $R_4$, the latter founds that the position (0,-1) is already taken. Thus, $R_4$ assumes that it can localize itself at (1,1). $R_4$ can localize itself at (1,1) only if the flag stored against (1,1) within the gene is 1 and one of its neighbours is in the \textit{Active} state. Since this is true in the case shown in Fig. \ref{trilat}, $R_4$ stores the associated \textit{NN} value from the gene into its memory and commences to emit its coordinates. 
         \begin{figure}[t]
             \centering
             \includegraphics[scale=.45]{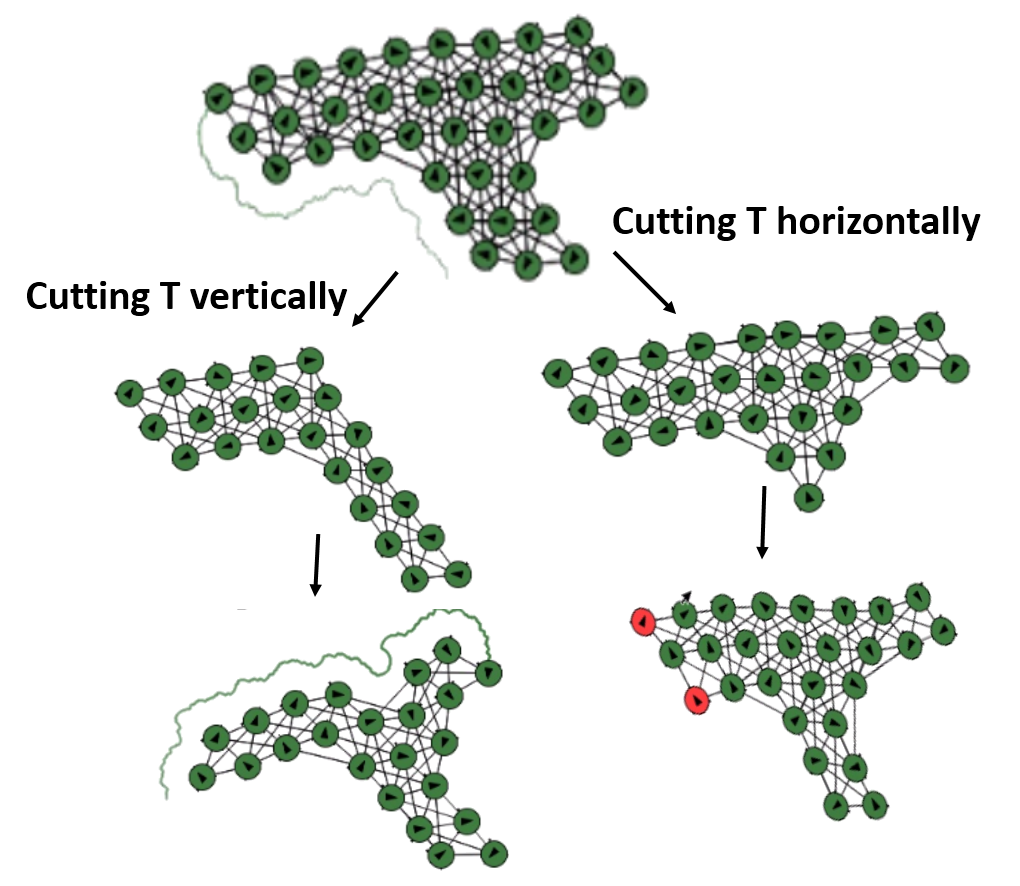}
             \caption{Regeneration of Shape `T' } 
            \label{result_3}
         \end{figure}
         
\begin{figure*}[t]
\minipage{0.45\textwidth}
  \includegraphics[width=\linewidth]{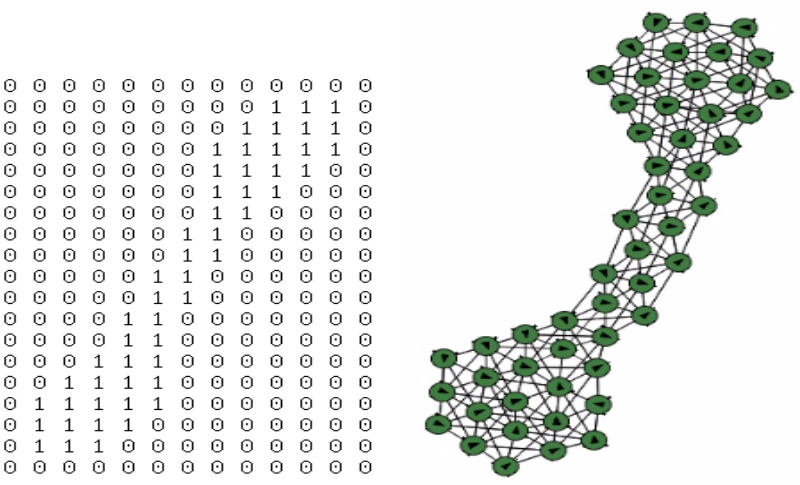}
  \caption{Dumbbell shape generation from the given bitmap image}\label{result_4}
\endminipage\hfill
\minipage{0.45\textwidth}
  \includegraphics[width=\linewidth]{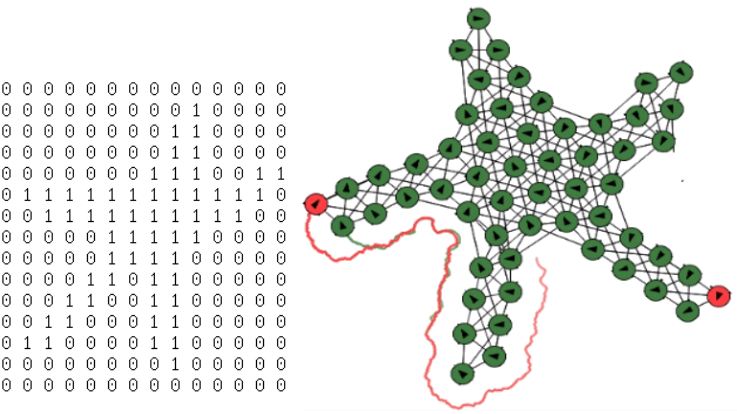}
  \caption{Starfish Shape generation from the given bitmap image}\label{result_5}
\endminipage
\end{figure*}

%
\section{Results}
To test the efficacy of the proposed algorithm, we used the Kilombo simulator. Kilombo is a C-based simulator developed to test swarm algorithms with Kilobots. The main reason behind the use of this simulator is that the code developed over Kilombo can be directly executed on an actual Kilobot. Kilobots are small 3.3 cm tall low-cost mobile robots with minimal computational capability. Instead of traditional wheels, they use vibration motors to effect sliding movements \cite{rubenstein2014programmable}. The robots can sense and communicate their distance from other robots using infrared light reflected off their surfaces. They can detect robots up to a distance of 7 cm. In addition, the robots can also broadcast messages to others.

Each Kilobot was initially provided with the input bitmap image of the desired shape to be formed and maintained. After feeding the input image, the Kilobots convert the bitmap into a gene. The remaining robots were organized in the form of a queue, and the robots were released one by one from the beginning of the queue. It may be noted that the remaining robots could be positioned in any disordered manner provided that they are minimally connected. The initial testing of the algorithm was performed on alphabetical shapes. Fig. \ref{result_1} shows the results generated when the simulated Kilobots were fed the bitmap of the alphabet `T'.

As can be seen from Fig. \ref{result_1}, the Kilobots cooperate and generate the shape of the alphabet `T' in a distributed manner. The line connecting any two Kilobots indicates that they are in the communication range of the other. The five differently coloured robots shown in Fig.  \ref{result_1}(a) indicate their respective states. Further, in the simulator, a moving Kilobot leaves behind a trail to show the path followed by it. A queue of Orange coloured robots in the \textit{Queued} state can also be seen in Fig.  \ref{result_1}(a),  \ref{result_1}(b) and  \ref{result_1}(c).
The Shape formation of the algorithm starts with the transition of the robots placed at the end of the queues from the \textit{Queued} state to the \textit{Search} state. Robots in the \textit{Search} state begin moving towards the already placed robots and place themselves near any robot in the \textit{Active} state and thus transit from the \textit{Search} to the \textit{Inactive} state. The Kilobots start moving from the end of the queue and place themselves due to which the shape continues to form as seen in later stages in Fig. \ref{result_1}. The proposed algorithm endeavours to be realistic by assuming that there can be losses in transmission. We have taken an error probability of 0.2 (loss in communication) which leads to minor deformities in the final shape formed. The results of a simulation run wherein, due to errors in message transmission, two of the robots (coloured violet and red) were not in a position to correctly sense the state transitions of their neighbours is shown in Fig. \ref{result_2}. This resulted in their inability to transit to the \textit{Stable} state even though the shape was formed. It may be noted that the resulting shape is not similar to the one observed in the bitmap. The bitmap is in a Cartesian coordinate system which gets skewed after adjusting in an SCS, as also shown in Fig. \ref{scs}. The robots then follow this skewed bitmap and arrange themselves in a triangular lattice form which finally results in a  skewed target shape.

\renewcommand{\thefootnote}{\faYoutubePlay}                       
The damage considered in this system consists of the cutting of the shape initially formed by the robots, thereby separating them into two parts. Each of these parts is separately capable of regenerating the original shape depending on the numbers of remaining robots in the respective parts. The regeneration of one such part is shown in Fig. \ref{result_3}. The shape was damaged by cutting it vertically (left branch in Fig. \ref{result_3}) and then cutting it horizontally (right branch). The Kilobots whose neighbours went missing, detect this damage and trigger the regeneration algorithm which in turn estimates the number of remaining Kilobots in the remnant to eventually reorganize to the new scaled down version of the original shape. During regeneration, the new shape can have an entirely new orientation as it depends on the coordinates of the initial three seed Kilobots' which mark the start of the shape regeneration algorithm. Other shapes that were considered include the dumbbell and starfish whose matrix representation and final shapes are shown in Fig. \ref{result_4} and \ref{result_5}. A video on the various shapes formed and regenerated by the algorithm can be found at the link\footnote[1]{\url{https://goo.gl/Qv7LMW}}.

The proposed algorithm is able to form and regenerate shapes with continuous surfaces. While filled shapes were formed without issues, those having unfilled and enclosed areas, such as the letter `O', seemed to be deformed.  Prima facie observations revealed that while the robots move to align the inside of an enclosed  unfilled shape, the open area gets enclosed and the robots get trapped within. Since we have introduced noise into the system, this entrapment of robot(s) could lead them to non-determinism. 
In addition when the robots are about to connect the enclosed portion of the shape, it may happen that some of them do not find enough space to enter the enclosure so as to complete the shape. This lack of space could make the stable robots to move away and avoid collision. This in turn can lead to an irregularity in the triangular lattice structure. If such stable robots move away beyond the communication range of their stable neighbours, the latter could detect this as a damage to the shape and initiate an undesired regeneration process. 
The problem could be solved if an additional state is introduced to categorize the robots in the periphery of the enclosed unfilled area so that further robots entering this area will detect this state and not venture to station themselves. This would mean a corresponding addition of a flag in the gene to indicate such enclosed and unfilled areas. 

\section{Conclusions}
Given a set of robots, the proposed algorithm is not only able to generate the given shape but also to ensure that any divisive damage caused to the shape can make each remnant to regenerate a scaled-down version of the same shape without external intervention. Damage could occur in the form of a robot or a set of robots malfunctioning or lose charge. All activities in the algorithm are performed in a  distributed and decentralized manner. The triangular lattice structure based on which the shape is formed helps identify the extent of loss of robots and also trigger the regeneration mechanism. The algorithm is novel because it handles both the generation and regeneration of a given shape without any intervention, much like biological cell division. The algorithm can be used in robotic swarms which need to divide and regroup in some given patterns. By providing, more than one image map, it may be possible for a leader robot to optimally decide the best shape the remaining set of robots should form. Such an algorithm could be useful in applications targeting exploration of inaccessible or risk prone areas. 
We are currently in the process of enhancing the algorithm so that the two scaled-down shapes can merge again to form the original shape. This feature would be useful when a swarm of robots travelling in a specific formation (shape) is split into two, due to an obstacle in their path, thus forming two scaled down versions. We are also working towards enhancing the algorithm to form and regenerate shapes with high complexity such as the ones with holes.

\bibliographystyle{ACM-Reference-Format}


\end{document}